\documentclass[letterpaper, 10 pt, journal, twoside]{IEEEtran}
\IEEEoverridecommandlockouts                              % This command is only needed if 
\usepackage{amsmath}
\usepackage{graphicx}
\graphicspath{ {./Pics/} }
\usepackage{enumerate}
\usepackage{times}
\usepackage{epsfig}
\usepackage{amssymb}
\usepackage{booktabs}
\usepackage{pifont}

\usepackage{bm}
\usepackage{comment}
\usepackage[dvipsnames]{xcolor}

\usepackage{enumitem}
\usepackage{booktabs}
\usepackage{array,multirow}

\usepackage{xspace}
\usepackage{booktabs} % for better-looking tables
\usepackage{graphicx} % for \scalebox

\usepackage{soul}

\newcommand{\ie}{\emph{i.e.}\xspace}
\newcommand{\eg}{\emph{e.g.}\xspace}

\newcommand{\etal}{\emph{et al.}\xspace}

\newcommand{\cmark}{\ding{51}}
\newcommand{\xmark}{\ding{55}}
\newcommand{\Fig}{Fig.\xspace}

\newcommand{\siminj}[1][black]{\textcolor{#1}}

\title{\LARGE \bf

 Real-time Trajectory-based Social Group Detection 
}
\author{Simindokht Jahangard, Munawar Hayat and Hamid Rezatofighi
\thanks{Simindokht Jahangard, Munavar Hayat and Hamid Rezatofighi are with Faculty of Information Technology, Monash University, Australia.
        {\tt\small simindokht.jahangard@monash.edu}}%
 }

\begin{document}

\maketitle
\thispagestyle{empty}
\pagestyle{empty}

%%%%%%%%%%%%%%%%%%%%%%%%%%%%%%%%%%%%%%%%%%%%%%%%%%%%%%%%%%%%%%%%%%%%%%%%%%%%%%%%
\begin{abstract}
Social group detection is a crucial aspect of various robotic applications, including robot navigation and human-robot interactions. To date, a range of model-based techniques have been employed to address this challenge, such as the F-formation and trajectory similarity frameworks. However, these approaches often fail to provide reliable results in crowded and dynamic scenarios. Recent advancements in this area have mainly focused on learning-based methods, such as deep neural networks that use visual content or human pose. Although visual content based methods have demonstrated promising performance on large-scale datasets, their computational complexity poses a significant barrier to their practical use in real-time applications. 
%Additionally, a recent self-supervised pose-based method has been found to be highly sensitive to noise, resulting in inaccurate or unstable representations. 
To address these issues, we propose a simple and efficient framework for social group detection. Our approach explores the impact of motion trajectory on social grouping and utilizes a novel, reliable, and fast data-driven method. We formulate the individuals in a scene as a graph, where the nodes are represented by LSTM-encoded trajectories and the edges are defined by the distances between each pair of tracks. Our framework employs a modified graph transformer module and graph clustering losses to detect social groups. Our experiments on the popular JRDB-Act dataset reveal noticeable improvements in performance, with relative improvements ranging from 2\% to 11\%. Furthermore, our framework is significantly faster, with up to 12x faster inference times compared to state-of-the-art methods under the same computation resources. These results demonstrate that our proposed method is suitable for real-time robotic applications.
\end{abstract}


\section{INTRODUCTION}

Detecting social groups has numerous applications in the field of robotics, including robot navigation~\cite{kruse2013human}, tele-operation robots, service robots, coworker robots~\cite{kubota2019activity,hanheide2017and}, tele-presence robots~\cite{barua2020can}, and autonomous driving cars~\cite{caba2015activitynet}. In these applications, accurately grouping individuals, interpreting and predicting human behavior, and responding them in a timely manner are critical tasks. A thorough understanding of social groups is essential for robots to effectively interact with people and analyze individual behavior in unconstrained environments. The development of a real-time system is imperative in these applications, as it allows the robot to perceive its environment and human behavior and interact with them in real-time.\\
% Detecting social groups has many applications in
% % surveillance~\cite{collins2000introduction} and
% robotics, \eg robot navigation~\cite{kruse2013human}, tele-operation robots, service robots, coworker robots~\cite{kubota2019activity,hanheide2017and}, tele-presence robots \cite{barua2020can} and autonomous driving cars~\cite{caba2015activitynet}. In these applications, correctly grouping individuals, interpreting, predicting and responding to human behaviors are essential tasks. An adequate understanding of social groups makes interacting with people feasible for robots and enables an accurate analysis of individual's behaviour in uncontrolled environments. In addition, developing a real-time system is imperative in these robotic applications, \eg, allowing the robot to perceive the environment and human behaviour and interact with them in real-time.\\
\begin{figure}[t]
\begin{center}
\scalebox{0.96}{
  \includegraphics[width=1\linewidth]{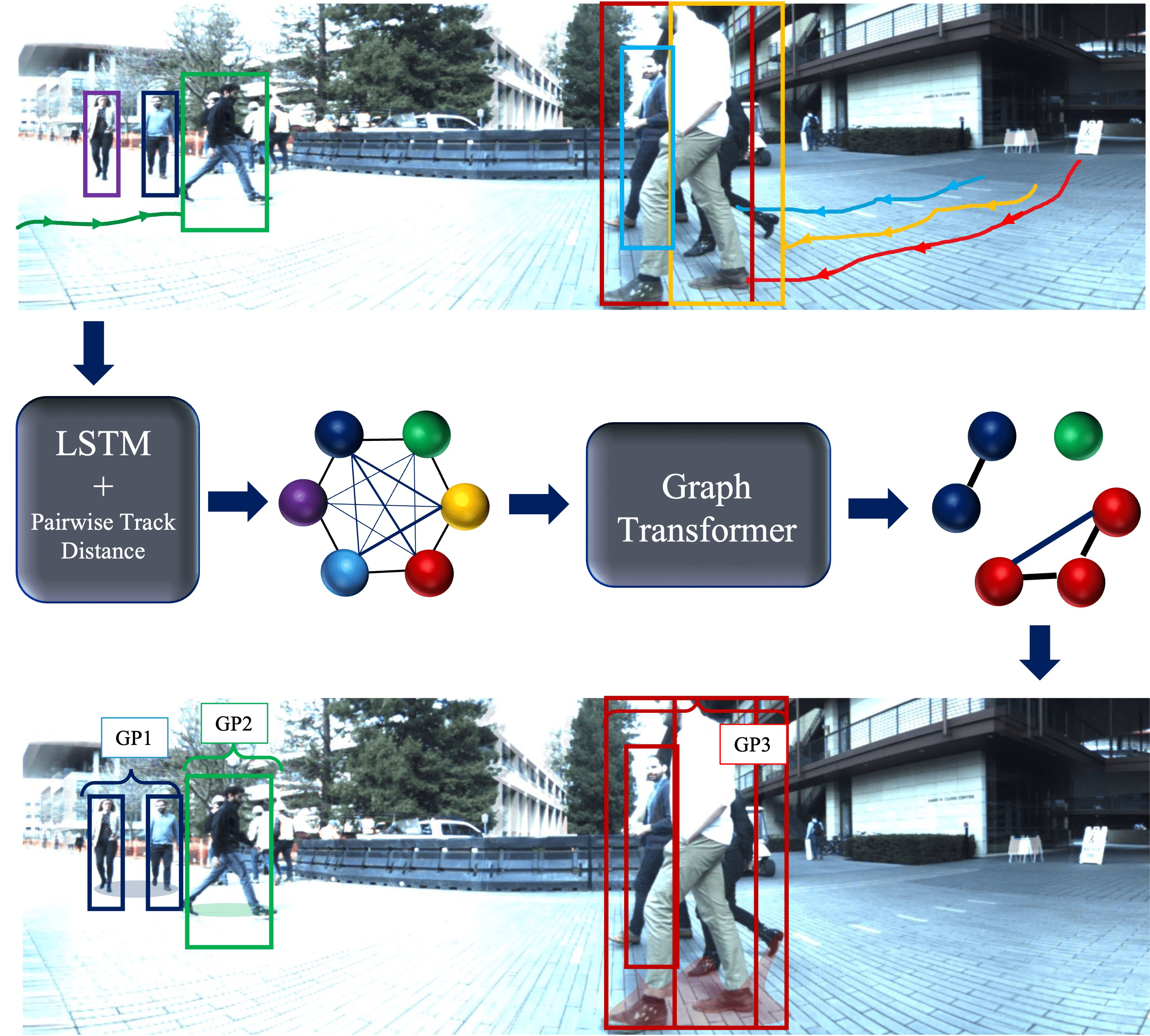}
}
\end{center}
    \caption{ A schematic representation of our proposed framework. A set of tracks, represented as LSTM encoded features, serve as input to our model. Using a pairwise distance metric between these trajectories, a graph is constructed and fed into a graph transformer. The graph transformer clusters the nodes in the graph, ultimately allowing for the prediction of social groups.}
\label{fig:Summary_Framework}
\end{figure}
% Social group detection also facilitates solving other problems from videos, such as semantic scene segmentation tracking \cite{liu2016multi}, anomaly detection \cite{palanisamy2017group}, crowd tracking and event recognition \cite{yuan2014online}.
% \hrt{we need a  concrete motivation, just focus on its robotics application (you can start having application in sureviellence and robotics, but later focus on robotic motivation) and highlight importance of being real-time process} \\
Grouping individuals in a crowd presents a significant challenge due to the complexity and unpredictability of human social behavior. The major obstacle in group detection is encoding group characteristics with compact and meaningful features that can form well-defined clusters.
To address this challenge, researchers have focused on grouping individuals based on their interactions~\cite{choi2014discovering,ge2012vision,hung2011detecting}, or by using F-formation techniques~\cite{kendon1990conducting} to model group formations~\cite{barua2020let, setti2015f}. Although these methods can effectively model group formations, they are limited to modelling only static groups~\cite{xie2019jointly}. Some efforts have been made to use trajectory information to detect groups based on similar temporal features~\cite{atev2010clustering, solera2015socially, chen2017anchor, ayazoglu2013finding}. However, these methods often struggle in challenging scenarios due to their limited generalization capability~\cite{li2021social}.
With the advent of deep learning, recent approaches have relied on spatio-temporal representations using 3D neural networks such as I3D ~\cite{ehsanpour2020joint,ehsanpour2022jrdb}.
These methods combine visual and geometrical features to detect social groups. \siminj {Additionally, a hierarchical graph neural network was proposed in~\cite{han2022panoramic} to model mutual social relations in a crowd of people. While these approaches have shown promising results, they are computationally expensive due to the dependency of their design in 3D neural networks, and, thus, may not be suitable for real-time applications in robotics. Recently~\cite{li2022self} proposed to model multi-person behavior relationships by leveraging a self-supervised pre-training strategy and developing a self-attention network for group detection. This framework assumes that body pose and global position of all the individuals are available as inputs. 
% \MH{the next sentence sounds very vague. can we be specific. exactly, what is the limitation of \cite{li2022self}}
Although the utilization of pose seems to be intuitive for better capturing human movements and interactions, in practice, the performance of the proposed method is susceptible to the pose estimation' noise and errors. In fact, the inherent ambiguity of pose creates difficulties in interpretation, especially in scenarios where multiple individuals are in close proximity. In addition, this framework uses a two-stage training process. However, the effectiveness of the social relation embedding network in the first stage significantly impacts the performance of the second stage. In other words, any deficiencies in the first stage could carry over and adversely affect the model's overall performance. This shortcoming is also validated by our experiments (Table ~\ref{table:ValidPerforme}).}
We introduce a simple, swift and efficient framework to tackle the challenges involved in detecting social groups, as depicted in Fig.\ref{fig:Summary_Framework}. Our approach exclusively employs bounding boxes as input, which is less prone to noise than other inputs such as body pose. Furthermore, we make a deliberate effort to create a lightweight model by using LSTM and graph transformer techniques, resulting in a model with only 700K parameters and capable of real-time performance. We represent each individual as a node within a graph structure, deriving the node's feature representation from encoding the subject's trajectory and the bounding box coordinates over time, utilizing an LSTM. We calculate inter-subject distances, based on the methodology described in~\cite{rezatofighi2019generalized}, and utilize them as edge features to serve as priors in the graph representation. We use a graph transformers to model the relationships between the nodes, where the node features encode the motion trajectory, and the edge features are based on the geometric distances between the subjects. The node features, updated through the graph transformer, are then combined with the geometric distances, and a multi-layer perception is employed to predict the number of social groups (cardinality) and track connectivity (adjacency matrix). \\
In this work, we present a novel and efficient deep architecture for detecting social groups. Our model has been evaluated on the large-scale and challenging JRDB dataset~\cite{ehsanpour2022jrdb} and has achieved state-of-the-art performance. The inference time of our model is under 0.034 seconds, which is significantly faster 
%(12 times faster) \MH{avoid repetition}
compared to recent works~\cite{ehsanpour2020joint,ehsanpour2022jrdb,han2022panoramic}, while utilizing the same computation resources.
The key contributions of this work are as follows:

\begin{enumerate}
\item A novel and computationally efficient neural architecture is proposed, which uses a combination of LSTM and a graph transformer, and only requires the subjects' trajectories as inputs.
\item A novel graph modeling approach is presented, where the motion of each trajectory is encoded as node features and the geometric distance between tracks is used as edge features.
\item Our model surpasses the performance of recent works and achieves a speed-up by up to 12 times.
\end{enumerate}

\section{Related works}
Social groups can be defined as emergent entities that are formed from the interactions and associations among two or more individuals who exhibit similar characteristics and exhibit a shared sense of unity. The automated detection of social groups is a topic of growing significance in the field of robotics, as it holds great promise for various applications, such as human-robot interaction, robot-assisted care, and social robot navigation. In this section, we present a concise overview of the recent developments in the research of social group detection.\\
% Social group is usually defined when two or more individuals interact with one another and share similar characteristics and a sense of unity. Automated detection of social groups finds its applications in robotics and has gained significant research traction lately. Here, we briefly review the most recent studies  on social group detection.\\
\textbf{F-formation technique:} This technique is a method that has been employed by several studies for modelling the spatial structure in social groups. Cristani \etal~\cite{cristani2011social} utilized statistical analysis to detect social interactions based on the spatial-orientational arrangements that are sociologically relevant. On the other hand, Setti \etal~\cite{setti2015f} utilized an efficient graph-cut based optimization to update the centers of the F-formations and prune unsupported groups using the Minimum Description Length principle. Despite its potential, the F-formation technique has certain limitations, as it is only capable of modelling the spatial structural formation and is not well-suited for dynamic group analysis~\cite{xie2019jointly}. This highlights the need for more advanced techniques that can model the temporal evolution of social groups, taking into account the changes in both the spatial and behavioral aspects of the group members.\\
% \textbf{F-formation technique:} Some studies employed F-formation technique to model the spatial structure in social group.
% Cristani \etal~\cite{cristani2011social} used statistical analysis of spatial-orientational arrangements having a sociological relevance to detect social interaction.
% Setti \etal~\cite{setti2015f} utilised efficient graph-cut
% based optimization and then update the centres of the F-formations followed by pruning unsupported groups in accordance with a Minimum
% Description Length. However, one of the downside of F-formation is that it is only able to model  the spatial structural formation and it is not suitable for dynamic group analysis ~\cite{xie2019jointly}.\\
\textbf{Trajectory-based:} Some studies have employed a trajectory-based approach. Pang \etal~\cite{pang2011detection} utilized a multivariate stochastic differential equation to model the position and velocity of individuals in real-time. Meanwhile, Atev\etal~\cite{atev2010clustering} developed a method for clustering vehicle trajectories based on the combination of two spectral clustering methods and a trajectory similarity measure based on the Hausdorff distance. Solera \etal~\cite{solera2015socially} proposed a tracklet clustering approach that utilized a defined distance function and four features to identify both the physical and social identity of pedestrians and to determine the presence of a shared goal. Chen \etal~\cite{chen2017anchor} introduced an Anchor-based Manifold Ranking (AMR) approach to examine the topological relationship of individuals and the global consistency of crowds in dense environments, accurately recognizing groups through a coherent merging strategy. Ayazoglu \etal~\cite{ayazoglu2013finding} tackled the problem of casual interaction detection using a directed graph topology and modelled casual connections between trajectories using Granger Causality theory. However, despite these efforts, current approaches are insufficient for identifying groups in dense crowds. 
% \MH{the last sentence should be revised, e.g., A self-supervised interaction modelling approach is proposed in \cite{li2021social}}... 
% \MH{we should finsih this para with the limitations of the existing approaches that have been discused.} 
 A recent study~\cite{li2022self} presented a new framework for human group detection that uses human body pose trajectories as inputs and utilizes a two-stage multi-head approach, leveraging self-supervised social relation feature embedding and a self-attention inspired network. In this study, the use of human body pose estimation can enhance the recognition of human movement and interaction. However, it is important to note that estimation errors and noise can negatively impact the performance of the method. \\
\textbf{ Visual-content based approaches:} In recent years, researchers have attempted to extract group-related information from visual images. Das \etal proposed the framework named Group-Sense, which identifies interacting groups based on acoustic context~\cite{das2018groupsense}. Another study combined orientation and context similarity in a parameter-free framework to cluster feature points~\cite{wang2018detecting} . Xie \etal modeled the interaction between individuals using a causality-induced hierarchical Bayesian model, where Granger causality was characterized by multiple features~\cite{xie2019jointly}. Li \etal proposed a method that profiles group properties using crowd trajectories and semantic information, then models multi-feature consistency and inconsistency in a unified graph clustering technique~\cite{li2021social}. Ehsanpour \etal and Han \etal utilized an I3D feature extractor, a self-attention module, and a graph attention module, where each individual's social interactions are encoded in a spatio-temporal feature map. Their model recognizes individuals' actions and social activities of each social group, relying on other tasks such as individual action and social group activity detection using visual features~\cite{ehsanpour2020joint,ehsanpour2022jrdb,han2022panoramic}. The disadvantage of these visual-content based methods is that they require large-sized networks, such as I3D, for feature extraction, hindering their real-time applicability. To address these challenges and limitations, a more efficient method for robust social group detection in dynamic human environments is proposed.
 \begin{figure*}[t]
\begin{center}
\scalebox{0.96}{
  \includegraphics[width=1\linewidth]{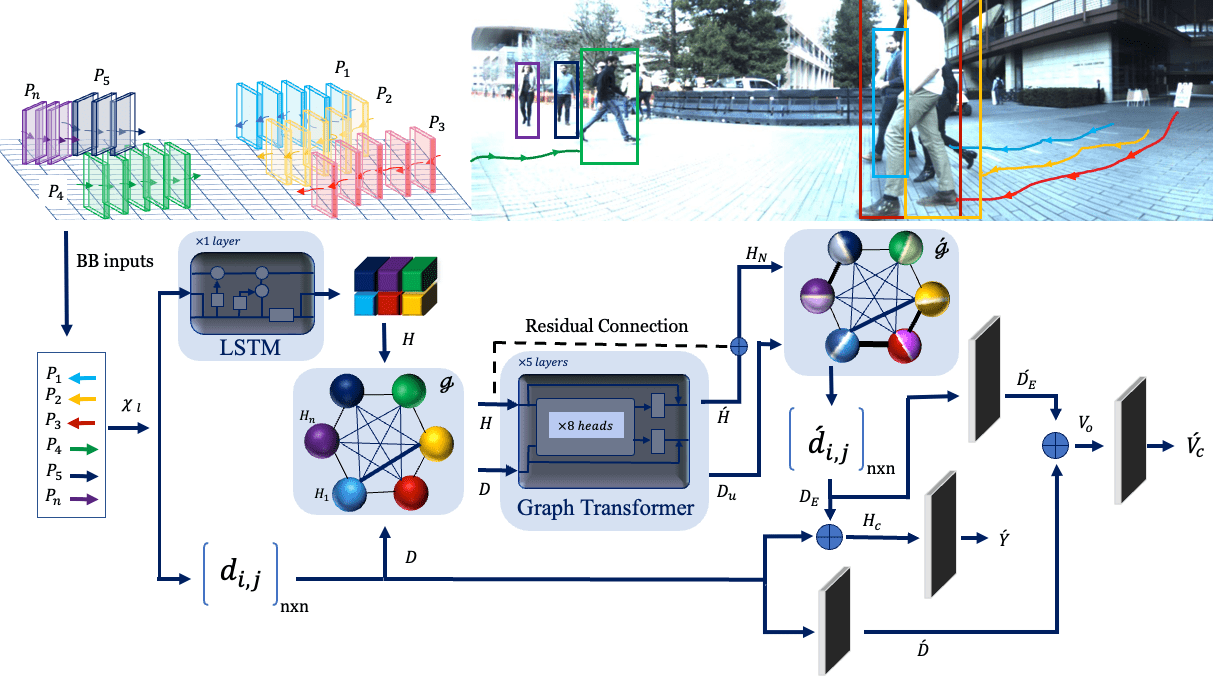}
}
\end{center}
    \caption{The overall pipeline of the proposed model consists of an LSTM encoder and a distance calculation module running in parallel, where $H$ and $D$ represent the node and edge of a graph, respectively. The generated graph is then processed by a modified graph transform module to update the node ($H_{N}$) and edge ($D_u$) features. The Euclidean distance between $H_{N}$ is calculated, concatenated with $D$, and fed into MLP layers to produce the adjacency matrix ($\acute{Y}$) and the number of groups ($\acute{V}_{c}$).  }
\label{fig:framework}
\end{figure*}
\section{Proposed Method}
The crucial aspect of social group detection lies in encoding a robust joint spatiotemporal feature representation that encompasses the movements and interactions of all individuals within a group. To accomplish this, our proposed method employs the use of Long Short-Term Memory (LSTM) to encode the individual's trajectories\footnote{Although, many other alternative temporal neural encoders, \eg Transformer and Temporal Convolutional Network (TCN), could also be considered for encoding our trajectories, we chose to use a shallow LSTM as a reasonable trade-off between performance and computational complexity.}. We then merge the encoded trajectories with a graph transformer to obtain a model with 700k parameters, a significant reduction compared to existing models. This approach leverages the notion that members within a group tend to exhibit similar behaviors and trajectories and combines this information with the inter-track distances to form a graph. The graph transformer learns a joint representation of this graph, which is then clustered into several connected components via graph clustering and spectral clustering to identify social groups. Fig.\ref{fig:framework} outlines the framework of our method, and the paper provides a comprehensive elaboration of its details. For the reader's ease of reference, Table ~\ref{table:Notations} summarizes the notations and symbols used throughout the paper.
% \hrt{This paragraph has many grammatical errors and most sentences should be revised (technically wrong) .} 
% In social group detection, encoding a proper joint spatio-temporal feature representation from all the individuals appearing in a group is crucial. \siminj{ To this end, 
% our approach first encodes individual's trajectories using Long Short-Term Memory (LSTM) to reduce computation burden. However, alternative methods such as Transformer and Temporal Convolutional Network (TCN) could also be used. We then combine this with a graph transformer, resulting in a model with 700k total parameters, which is significantly less compared to existing models. This benefits from the assumption that members in a group typically have similar trajectories and combines them with the distance between each pair of tracks} and combines them with a distance between each pair of tracks to generate a graph. Using a graph transformer, we next learn a joint representation for this built graph which is finally clustered into several connected components using a graph clustering spectral clustering in order to detect the social groups.  
%  \Fig~\ref{fig:framework} shows the pipeline of our framework. Here, we elaborate more on details of our method. \siminj{For the reader's convenience, a summary of the notations and symbols used in this paper is provided in Table 
\begin{table}[tbh]
  \begin{center}
    \scalebox{0.8}{
\begin{tabular}{c|l}
\toprule
Notations & Descriptions \\
\midrule
$\mathcal{X}_{\ell}$ & a set of different variable-length tracks \\
\midrule
${Y}_{{\ell}}$ & adjacency matrix between the tracks\\
\midrule
$P_{i}$ & $i_{th}$ instance track \\
\midrule
$P^{t_0:t_T}_i$ & sequence of i-th instance bounding boxes (track) over time span $t_0$-$t_T$ \\
\midrule
$B=F^{C}(A;\Theta)$ &  function named C with A as input, B as output and $\Theta$ as its parameter\\
\midrule
$d^{GIoU}_{i,j}$ &  a normalized GIoU distance between $i_{th}$ and $j_{th}$ instance \\
\midrule
$\tilde{d}_{0}(\{P^t_{i}\},\{P^t_{j}\})$ &  distance between $i_{th}$ and  $j_{th}$ instance\\
\midrule
 $D_{P_{i}}$ & time duration when the tracks $i$ exist \\
 \midrule
 $\bm{d}_{i,j}(P_{i},P_{j})$ & distance between two variable-length trajectories, $P_{i}$ and $P_{j}$\\
\midrule
$\bigwedge$ , $\bigoplus$  & logical operations functions\\
\midrule
$\mathbin\Vert$ & concatenation\\
\bottomrule
\end{tabular}}
\end{center}
\caption{Summary of Notation }
\label{table:Notations}
\end{table} 

\begin{figure}[t]
\begin{center}
\scalebox{0.99}{
  \includegraphics[width=1\linewidth]{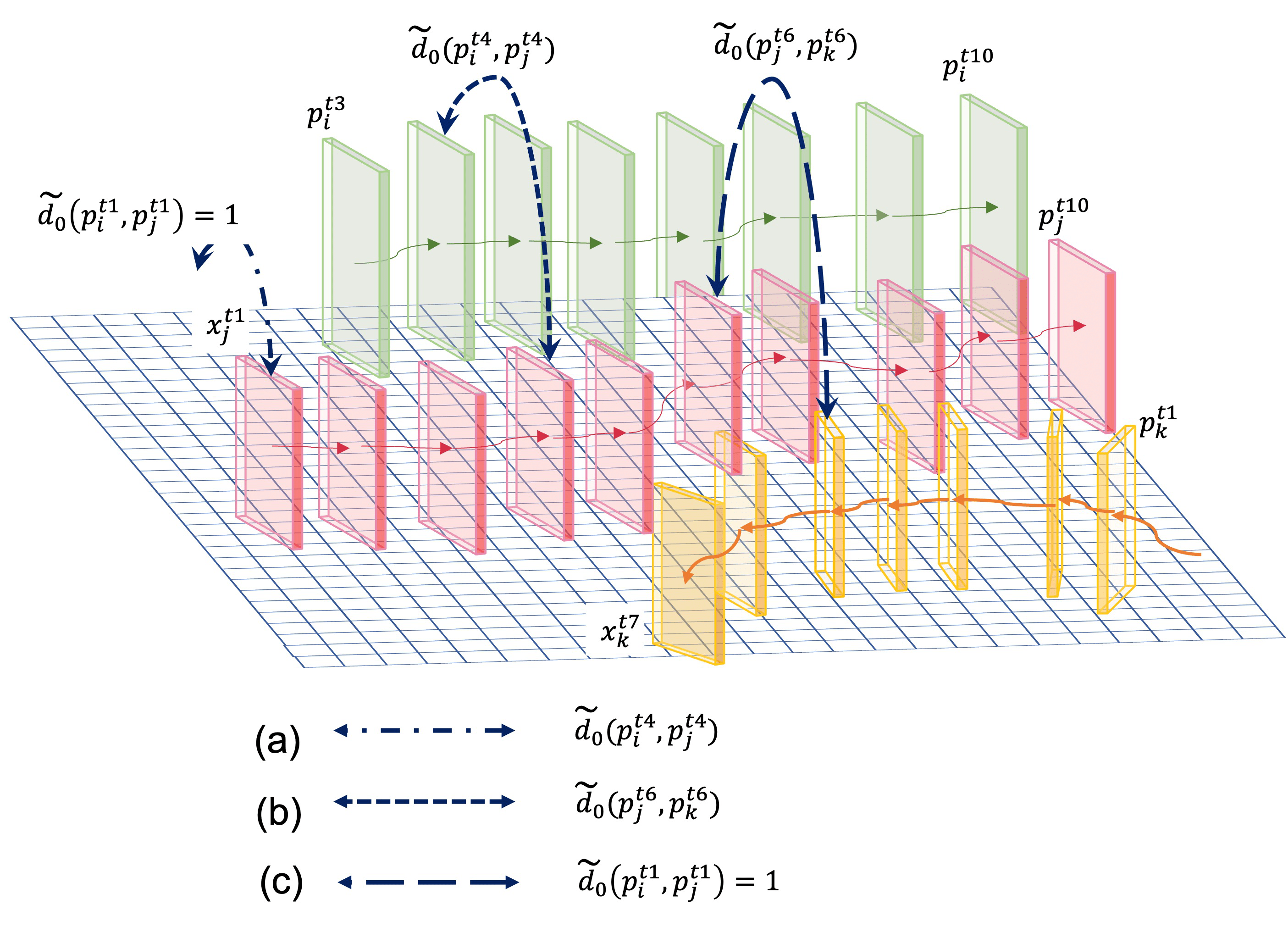}

}
\end{center}
    \caption{ Distance calculation among motion trajectories using GIoU metric based on Eq.(\ref{GIOU}). The scenarios depicted in (a) and (b) represent instances where both instances are present, while scenario (c) represents a situation where one of the instances is absent in the frame.}
\label{fig:GIOU}
\end{figure}

\textbf{Input and Output:} We train our model using a training set with $L$ samples, \ie $\{ \mathcal{X}_{\ell},{Y}_{{\ell}} \}_{1}^L$, where the input, $\mathcal{X}_{\ell}=\{P_{1}, P_{2},..., P_{n}\}$ is a set of different variable-length tracks (each track represents a single instance and consists of a sequence of bounding boxes). The output
 ${Y}_{{\ell}}=\bigg[y_{i,j}\bigg]_{n\times n} \in \mathbb{B}^{n\times n},y_{i,j} \in \{0,1 \}$, represents the adjacency matrix between the tracks. The value $y_{i,j}$ indicates if two track instances belong to the same group ($y_{i,j}=1$) or not $(y_{i,j}=0)$. \\
 % such that if two track instances are connected (belong to the same group), the corresponding value in this matrix is 1; otherwise is zero. \\
 %It defined as ${Y}_{{\ell}}=\Bigr[ y_{i,j} \Bigr]_{n\times n} \in \mathbb{B}^{n\times n},\vspace{}y_{i,j} \in \{0,1 \}$.
% \begin{equation}
% \mathcal{X}_{\ell}=\{P_1,P_2,...,P_n\} 
% \end{equation}
%  \begin{equation}
% Y_{\ell}=\Bigr[ y_{i,j} \Bigr]_{1}^L \in \mathbb{R},\vspace{}y_{i,j} \in \{0,1 \}
% \end{equation}
% $\mathcal{X}_{\ell}$ is a sequence of different tracks  which varies from 1 to $L$. 
The number of tracks is indicated by  $n$ and $P^{t_0:t_T}_i$ represents a track for instance $i$, which is a sequence of its track states, $P^{t_0:t_T}_i=( \mathrm{p}^{t_0}_{i},\cdots,\mathrm{p}^{t_T}_{i})$, \eg bounding box coordinates and sizes, within the time range from $t_0$ to $t_T$.\\
% $P^{t_0:t_T}_i=( \mathrm{p}^{t_0}_{i},\cdots,\mathrm{p}^{t_T}_{i})$, \eg, bounding box coordinates,   over its time span between $t_0$ and $t_T$.
% \begin{equation}
%  P_i=( \mathrm{x}^{t_0}_{i},\cdots,\mathrm{x}^{t_T}_{i})  
% \end{equation}
% in which, $\mathrm{p}^{t}_{i}$ 
% % (\mathrm{x}^{t_0}_{i_1_l},\mathrm{y}^{t_0}_{i_1_l},\mathrm{x}^{t_0}_{i_2_l},\mathrm{y}^{t_0}_{i_2_l})
% shows
% the coordination of bounding boxes
% that  $(\mathrm{x}^{t_0}_{i_1_l},\mathrm{y}^{t_0}_{i_1_l})$ defines coordinations of top left and $(\mathrm{x}^{t_0}_{i_2_l},\mathrm{y}^{t_0}_{i_2_l})$ down right of a bounding box 
% at time ${t}$.
\textbf{Spatio-Temporal Encoding:} 
% We consider 30 frames before the key-frame and choose every other frame. Therefore,
Each sequence of bounding boxes, $P^{t_0:t_T}_i$, is input to an LSTM module to extract the spatiotemporal characteristics of each instance's trajectory motion, as described by Eq. (\ref{Eq1}).
% Each sequence of bounding boxes, $P^{t_0:t_T}_i$ is fed to LSTM module to encode the spatio-temporal feature of each instance's trajectory motion Eq (1).
 \begin{equation}\label{Eq1}
H_{i}=F^{LSTM}( P^{t_0:t_T}_i;\Theta_{1}),  
\end{equation}
 where $F^{LSTM}( \cdot;\Theta_{1})$ is LSTM module that gets sequences of bounding boxes of a track and extracts its motion features and $H_{i}\in \mathbb{R}^{1\times N}$ indicates the encoded spatio-temporal feature vector for track $i$ as the output.\\
\textbf{Calculating a Distance Between Two Variable-length Tracks:} 
To determine a distance between two dynamic trajectories, $P_{i}$ and $P_{j}$,  the time-averaged distance from OSPA$^{(2)}$ (an extension of Optimal Sub-Pattern Assignment (OSPA))~\cite{beard2020solution} is utilized:
\begin{equation}\label{Eq2}
 \bm{d}_{i,j}(P_{i},P_{j})=\sum_{t\in P_{i} \cup P_{j}} \frac{\tilde{d}_{0}(\{P_{i}^t\},\{P_{j}^t\})}{\vert D_{P_{i}} \cup D_{P_{j}} \vert},
\end{equation}
%  \begin{equation}
%   P_{i}=( \mathrm{x}^{t_{\tau_a}}_{i},\cdots,\mathrm{x}^{t_{\tau_z}}_{i}), P_{j}^t=( \mathrm{x}^{t_{\acute{{\tau}_a}}}_{j},\cdots,\mathrm{x}^{t_{\acute{{\tau}_z}}}_{j}) 
% \end{equation}\\
where $D_{P_{i}}$ and $D_{P_{j}}$ represent the time duration when tracks $i$ and $j$ exist, respectively. When at least one of the tracks  $i$ and $j$  exists, the time duration is represented by 
$ D_{P_{i}}\cup D_{P_{j}}\neq \emptyset $.  The value of 
$\bm{d}_{i,j}(P_{i},P_{j}) $ is 0 if the time duration when the tracks $i$ and $j$ exist is an empty set, i.e.,  $ D_{P_{i}} \cup D_{P_{j}} = \emptyset $. \\
% where $D_{P_{i}}$ and $D_{P_{j}}$ are the time duration when the tracks $i$ and $j$ exist, respectively. Therefore, $ D_{P_{i}∪} \cup D_{P_{j}}  \neq \emptyset $ represent time duration when at least one of the track $i$ and $j$ exists and $\bmath{d}_{i,j}(P_{i},P_{j}) = 0$, if $ D_{P_{i}} \cup D_{P_{j}} = \emptyset $,\\ 
% and $t_{\tau_z}>t_{\tau_a}}$ and $t_{\acute{{\tau}_z}}>t_{\acute{{\tau}_a}}}$, $t_{\tau_z},t_{\tau_a}},  t_{\acute{{\tau}_z}}$ , and $t_{\acute{{\tau}_a}}}$ varies from 0 to 15.
% If $\vert D_{P_{i}∪} \cup D_{P_{j}} \vert \neq \phi $ and $\b{d}(P_{i},P_{j}) = 0$, if $\vert D_{P_{i}^{t} ∪} \cup D_{P_{j}^t} \vert = \phi $.\\
Additionally, $\tilde{d}_{0}(\{P^t_{i}\},\{P^t_{j}\})$ calculates the distance between two Singleton sets, and is determined using equations inspired by the OSPA set distance formula ~\cite{rezatofighi2020trustworthy} ~\cite{schuhmacher2008consistent} :  
%each pair of bounding boxes $(\mathrm{p}_{i},\mathrm{p}_{j})$ in a track sequence by utilising the normalised GIoU \cite{rezatofighi2019generalized} and it is calculated as below:
% \begin{equation}
% d_{0}(\mathrm{x}^{t}_{i},\mathrm{x}^{t}_{j})=\frac{1}{N}\left(min_{\pi\in\Pi_{N}}\sum_{i=1}^M \b{d}(\mathrm{x}^{t}_{i},\mathrm{x}^{t}_{\pi(j)})+ (N-M)\right)
% \end{equation}
% \begin{equation}
% \tilde{d}_{0}(\{P_{i}\},\{P_{j}\})=\frac{1}{N}\left(\sum_{i=1}^M d(\mathrm{p}^{t}_{i},\mathrm{p}^{t}_{j})+ (N-M)\right)
% \end{equation}
%  \begin{equation}
% D_{\ell=\Bigr[ d_{i,j} \Bigr]_{n\times n} \in \mathbb{R},\vspace{}d_{i,j} \in [\ 0-1]\ 
% \end{equation}
\begin{equation} \label{GIOU}
\tilde{d}_{0}(\{P^t_{i}\},\{P^t_{j}\}) = \begin{cases}
d^{GIoU}_{i,j} & \text{if $\vert\{P^t_{i}\}\vert\wedge\vert\{P^t_{j}\}\vert = 1$},\\
1 & \text{if $\vert\{P^t_{i}\}\vert\oplus\vert\{P^t_{j}\}\vert =  1$},\\
0 & \text{Otherwise},
\end{cases}
\end{equation}
% It is supposed that $ N=M=1$ and if $d(\mathrm{x}^{t}_{i},\mathrm{x}^{t}_{j})=1$, if one of them is empty. additionally: $d(\mathrm{x}^{t}_{i},\mathrm{x}^{t}_{j})=d(\mathrm{x}^{t}_{j},\mathrm{x}^{t}_{i})$.
where $d^{GIoU}_{i,j} = \frac{1-GIoU(i,j)}{2}$ is used to calculate the normalized GIoU distance between tracks $i$ and $j$, with a range of values between 0 and 1, when both tracks exist at time $t$, \ie $\vert\{P^t_{i}\}\vert\wedge\vert\{P^t_{j}\}\vert = 1$.  If at least one instance does not exist (Second term) i.e., $\vert\{P^t_{i}\}\vert\oplus\vert\{P^t_{i}\}\vert = 1$ , the distance is set to the maximum value of 1.
% In this case, the distance is 1 which is the maximum value of $d^{GIoU}_{i,j}$.
Otherwise, the distance is 0, as shown in third term. The calculation process is demonstrated in \Fig~\ref{fig:GIOU}. \\
\textbf{Modeling Graph:} 
The graph $\mathcal{G} = \{H , D\}$ models the relationship between tracks (instances). The set $H$ contains nodes that represent the spatiotemporal features of each track (encoded by an LSTM), while the set $D$ consists of edges that denote the distances between tracks, defined above. The graph transformer, as described in~\cite{dwivedi2020Generalization}, is employed to modify and enhance the features of both nodes and edges.

% The graph $\mathcal{G} = \{H , D\}$ represents the relationship between tracks (instances), where $H$ is a set of nodes that denote the spatiotemporal features of tracks, and $D$ is a set of edges that signifies the distance between tracks.  The graph transformer  \cite{dwivedi2020generalization} is utilized to modify and update the node and edge features.
% % Considering the graph 
% $\mathcal{G} = \{H , D\}$, where $H$ is the node set denoting spatio-temporal features of tracks and $D$ is the edge set that represents the distance between tracks.
% We use graph transformer \cite{dwivedi2020generalization} to update node and edge features. 
% More details about used graph transformer are illustrated in \Fig~\ref{fig:Graph transformer}.
 \begin{equation}
\acute{\mathcal{G}}=F^{GR}( \mathcal{G};\Theta_{2}),
\end{equation}

Where $F^{GR}( \cdot;\Theta_{2})$ is graph transformer module that processes generated graph, $\mathcal{G} = \{H , D\}$, and produces updated graph   $\acute{\mathcal{G}}=\{H_{N},D_{u}\}$, $H_{N}$  denoted the updated node features that defined as $H_{N}= H + \acute{H}$ and $\acute{H}$ shows the residual connection, illustrated in \Fig~\ref{fig:framework}. Also, $D_{u}$ is updated edge features.\\ 
Then the Euclidean distance, $D_{E}$, between each node feature, $H_{N}$, has been calculated and concatenated
with the distance between tracks.
\begin{equation} \label{De}
D_{E}=\Bigr[\acute{d}_{i,j}\Bigr]_{n\times n} \in \mathbb{R},\acute{d}_{i,j} \in [\ 0-1],
\end{equation}
\begin{equation} \label{Eq6}
H_{c}=D_{E}\mathbin\Vert D,
\end{equation}\\
where $\mathbin\Vert$ means concatenation.
 Finally, a simple module  that consists of MLPs (Multi-Layer Perceptrons) is utilized to predict the adjacency matrix, $\acute{Y}$. \\
\begin{equation} \label{Eq7}
\acute{Y}=F^{L_1}( H_{c} ; \Theta_{3}).  
\end{equation}
$F^{L_1}( \cdot; \Theta_{3})$ is an MLP layer,
$D_{E}$ and $D$ are passed through few MLP layers $\acute{D}_{E}=F^{L_2}( H_N ;\Theta_{4})  $ and $\acute{D}=F^{L_3}( D ;\Theta_{5})  $ then after concatenation ${V}_{o}=\acute{D}_{E}\mathbin\Vert \acute{D}$ feeds to a final linear layer to predict the number of social groups (cardinality), $\acute{V}_{c}$, \ie, 
\begin{equation} \label{Eq8}
\acute{V}_{c}=F^{L_4}( {V}_{o} ;\Theta_{6}),  
\end{equation}
% \hrt{what is $\acute{V}_{c}$ define it clearly. define all the neural functions }
where $F^{L_4}( \cdot ;\Theta_{6}) $ is an MLP layer, given concatenated vector ${V}_{o}$ and produces $\acute{V}_{c}$ which shows number of groups.\\
\textbf{Loss Function:} We used the loss function same as \cite{ehsanpour2022jrdb}, the loss is combination of three losses: $\mathcal{L}_{BCE}$, $\mathcal{L}_{eig}$ and $\mathcal{L}_{MSE}$, shown in Eq. (\ref{losses_eq}).
\begin{equation} \label{losses_eq}
\mathcal{L}_G=\lambda_1\mathcal{L}_{BCE}(\acute{Y},Y)\!+\!\lambda_2\mathcal{L}_{eig}(\acute{L},L)\!+\!  \lambda_3\mathcal{L}_{MSE}(\acute{V}_{c},V_{c}),
\end{equation}
% \begin{figure}[t]
% \begin{center}
% \scalebox{0.85}{
%   \includegraphics[width=1\linewidth]{Pics/graph transfor.jpg} 
% }
% \end{center}
%     \caption{ The illustration of graph transformer layers with edge features}
% \label{fig:Graph transformer}
% \end{figure}
% \hrt{from here onward, it requires revision}\\
where $\lambda_1$, $\lambda_2$ and $\lambda_3$ are the coefficient weights of the loss terms. The variables
with $\acute{(.)}$ denote the corresponding ground-truth labels.
\\
A binary cross-entropy loss, $\mathcal{L}_{BCE}$, is used between the elements of predicted adjacency, $\acute{Y}$, and ground truth adjacency, $Y$.  We also utilize $\mathcal{L}_{eig}$ denoted by Eq. (\ref{l_eig}) and is inspired by the fully differentiable, eigen decomposition-free loss proposed in~\cite{dang2018eigendecomposition} to train a deep network whose loss depends on the eigenvector of the matrix predicted by the network containing a single zero eigenvalue. The number of social groups (connected components) in ground truth matrix $Y$ corresponds to the number of zero  eigenvalues of its laplacian matrix $L$,
we calculate the laplacin matrix of $\acute{Y}$ noted by $\acute{L}$ to achieve zero eigenvalues, same as in $L$. To this end, Eq. (\ref{l_eig}) is utilized below. \\
% \hrt{the notation is confusing and inconsistent}
\begin{equation} \label{l_eig}
L_{eig}=e^{T}\acute{L}^{T}\acute{L}e +\alpha \exp(-\beta tr(\bar{\acute{L}}^T\bar{\acute{L}})),
\end{equation}
with $e$ as the ground truth eigenvector corresponding to zero eigenvalue,
the laplacian matrix $\acute{L}$ corresponds to the predicted similarity matrix $\acute{Y}$, while $\alpha$ and $\beta$ are coefficients. More details and the proof of Eq. (\ref{l_eig}) is stated in~\cite{ehsanpour2022jrdb}. 
\\
Finally, $\mathcal{L}_{MSE}$ is a mean square error (cardinality loss), which is used to predict the number of social groups (number of connected components).  \\
% \begin{table}[tbh]
%   \begin{center}
%     % {\small{
%     \scalebox{0.8}{
% \begin{tabular}{c|l}
% \toprule
% Notations & Descriptions \\
% \midrule
% $\mathcal{X}_{\ell}$ & a set of different variable size tracks \\
% \midrule
% ${Y}_{{\ell}}$ & adjacency matrix between the track\\
% \midrule
% $P_{i}$ & $i_{th}$ instance track \\
% \midrule
% $P^{t_0:t_T}_i$ &  i-th instance track over its time span between $t_0$ and $t_T$ \\
% \midrule
% $B=F^{C}(A;\Theta)$ &  Function named C with A as input, B as output and $\Theta$ as its parameter\\
% \midrule
% $d^{GIoU}_{i,j}$ &  a normalized GIoU distance between $i_{th}$ and $j_{th}$ instance \\
% \midrule

% $\tilde{d}_{0}(\{P^t_{i}\},\{P^t_{j}\})$ &  distance between $i_{th}$ and  $j_{th}$ instance\\
% \midrule

%  $D_{P_{i}}$ & time duration when the tracks $i$ exist \\
%  \midrule
%  $\bmath{d}_{i,j}(P_{i},P_{j})$ & distance between two variable-sized trajectories, $P_{i}$ and $P_{j}$\\
% \midrule
% $\bigwedge$ , $\bigoplus$  & logical operations functions\\
% \midrule
% $\mathbin\Vert$ & concatenation\\

% \bottomrule
% \end{tabular}
% }
% \end{center}
% \caption{Summary of Notation }
% \label{table:testPerforme}
% \end{table}

\textbf{Training:} Considering the compounded parameters of the suggested neural models as $\mathbf{\Theta} =( \Theta_{1}, \Theta_{2}, \dots,\Theta_{6})$, we train them end-to-end by minimizing the total loss over $\mathbf{\Theta}$, \ie $\mathbf{\Theta}^*=\arg\min_\mathbf{\Theta}\sum_{\ell = 1}^L\mathcal{L}^{\ell}_G$, using stochastic gradient descent~\cite{bottou2012stochastic}. \\
% \begin{equation} \label{minimization of total}
% \mathbf{\Theta}^*=\arg\min_\mathbf{\Theta}\sum_{\ell = 1}^L\mathcal{L}^{\ell}_G 
% \end{equation}
\textbf{Inference:}
We use graph spectral clustering~\cite{zelnik2004self} on the predicted adjacency matrix, $\acute{Y}$, and the predicted number of social groups, $\acute{V}_{c}$, to detect the social groups. Spectral clustering employs the heuristic~\cite{ng2001spectral} to determine the number of social groups present.

\section{Experiments}
\begin{figure*}
\begin{center}
\scalebox{0.96}{
  \includegraphics[width=1\linewidth]{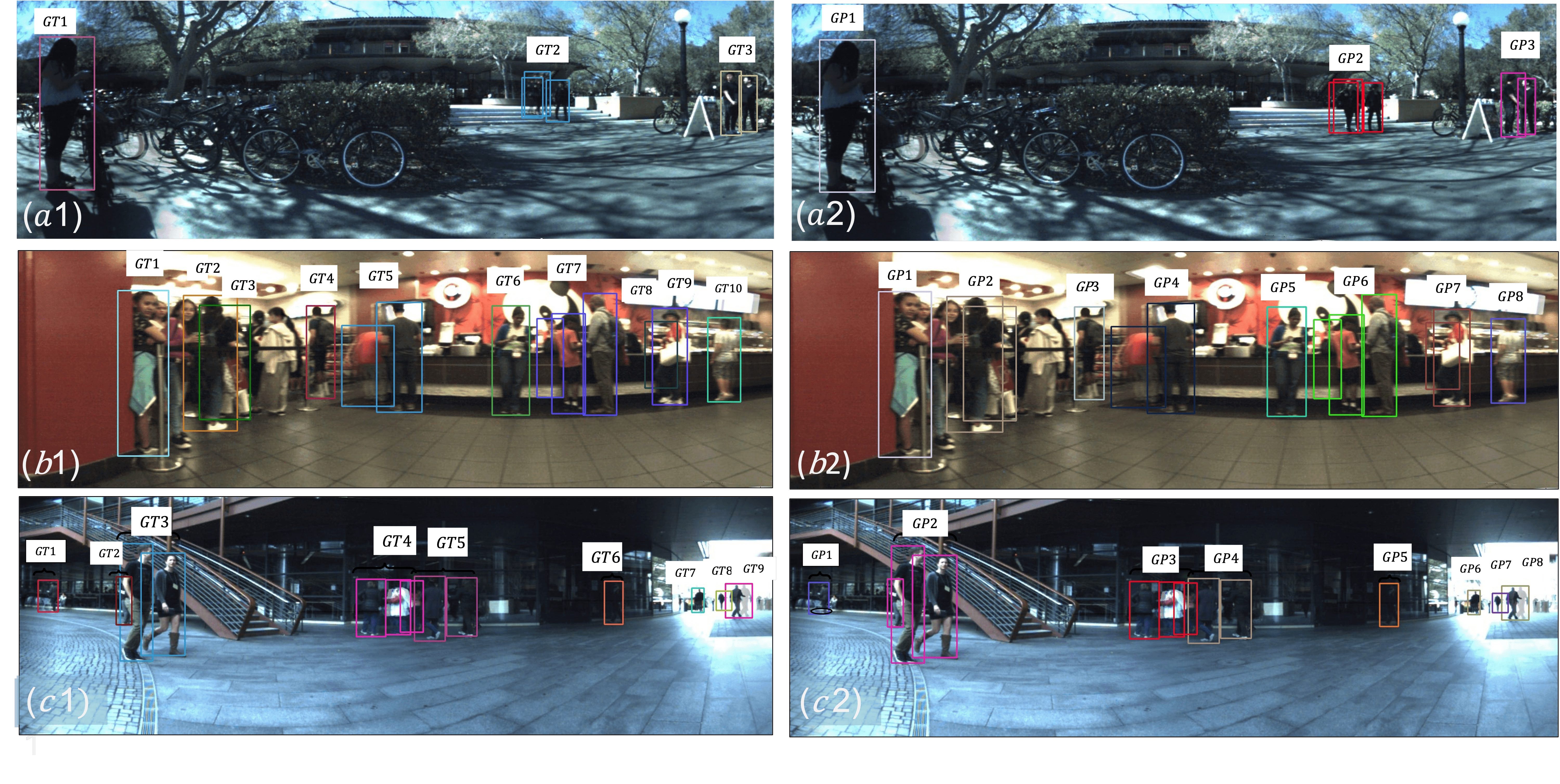} 
}
\end{center}\vspace{-2em}
    \caption{The visualization of our model's performance on the validation set is depicted in $(a1),(b1)$, and $(c1)$, which represent the ground truth, while $(a2),(b2)$, and $(c2)$ depict the groups detected by our model. Our model demonstrates exceptional results in both straightforward $(a2)$ and complex $(b2)$ and $(c2)$ scenes. However, there is still scope for improvement. For instance, in $(b2)$, the group "GP2" is not fully associated with the ground truth groups "GT2" and "GT3" in $(b1)$. }
\label{fig:vis1}
\end{figure*}

% \begin{figure*}[t]
% \begin{center}
% \scalebox{0.99}{
%   \includegraphics[width=1\linewidth]{Pics/C_D.jpg}
% }
% \end{center}
%     \caption{The visualization on validation set $
%     (c)$ and $(d)$ show the ground truth and $
%     (\acute{c})$ and $(\acute{d})$ illustrate the detected groups.}
% \label{fig:vis2}
% \end{figure*}
% \begin{figure*}[t]
% \begin{center}
% \scalebox{0.99}{
%   \includegraphics[width=1\linewidth]{Pics/sample4.jpg}
% }
% \end{center}
%     \caption{ visualization_GT}
% \label{fig:framework}
% \end{figure*}

% \begin{figure*}[t]
% \begin{center}
% \scalebox{0.99}{
%   \includegraphics[width=1\linewidth]{Pics/sample2.jpg}
% }
% \end{center}
%     \caption{ visualization_Pre*************************************************************************************************************************************************}
% \label{fig:framework}
% \end{figure*}

\begin{table*}[tbh]
  \begin{center}
    {\small{
\begin{tabular}{c|c|ccccc|c}
\toprule
Set&Method &G1 AP$\uparrow$ & G2 AP$\uparrow$ & G3 AP$\uparrow$ & G4 AP$\uparrow$ & G5$^+$ AP$\uparrow$ & mAP$\uparrow$  \\
\midrule
            &\siminj{Li \etal}~\cite{li2022self}&3.1 & 25.0 & 17.5 & 45.6 & 25.2 & 23.3\\
            &\siminj{Ehsanpour \etal}~\cite{ehsanpour2020joint}&8.0 & 29.3& 37.5 & 65.4 & 67.0 & 41.4 \\
Validation  &\siminj{Han \etal}~\cite{ han2022panoramic}&52.0 & 59.2 & 46.7 & 46.6 & 31.1 & 47.1 \\
            &\siminj{Ehsanpour \etal}~\cite{ehsanpour2022jrdb}&81.4 & 64.8 & 49.1 & 63.2 & 37.2 & 59.2 \\
            &{\bf Ours} &71.9 & 73.6& 64.0&71.2&48.6&\bf 65.9\\
            \midrule
            & \siminj{Ehsanpour \etal}~\cite{ehsanpour2020joint} & 11.2	& 24.6 &	21.8 &	41.2&	23.5 & 24.5 \\
 Test           & \siminj{Ehsanpour \etal}~\cite{ehsanpour2022jrdb}& 56.3 &	39.4&	24.3&	22.9&	15.3 &31.6 \\
            & {\bf Ours} & 19.5&39.7 &30&48.6&24.7&\bf 32.5\\
\bottomrule
\end{tabular}
}}
\end{center}
\caption{Social Group Detection on JRDB-Act (Validation/Test)-Set using the ground truth detection/tracks.}
\label{table:ValidPerforme}
\end{table*}
\begin{table}[tbh]
  \begin{center}
    {\small{
\begin{tabular}{l||cc}
\toprule
Method & Num. Parameters$\downarrow$ & fps$\uparrow$  \\
\midrule
Ehsanpour \etal~\cite{ehsanpour2020joint}&27M$^{+}$& 2.40$^{-}$ \\
Ehsanpour \etal~\cite{ehsanpour2022jrdb}& 27M& 2.40\\
Han \etal~\cite{ han2022panoramic}&23M$^{+}$& ~2.3$^{-}$ \\
Li \etal~\cite{li2022self}&1.6M& 29.41$^{-}$ \\
{\bf Ours}& 700K & 29.41 \\
% \siminj{Ehsanpour \etal}~\cite{ehsanpour2020joint} & 11.2	& 24.6 &	21.8 &	41.2&	23.5 & 24.5 \\
% \siminj{Ehsanpour \etal}~\cite{ehsanpour2022jrdb}& 56.3 &	39.4&	24.3&	22.9&	15.3 &31.6
% \\
% {\bf Ours} & 19.5&39.7 &30&48.6&24.7&\bf 32.5\\
\bottomrule
\end{tabular}
}}
\end{center}
\caption{Comparison Between Number of Parameters and Inference FPS. "$^{+}$" and "$^{-}$" indicate a greater and a lower values, respectively. } 
\label{table:testPerforme}
\end{table}
\textbf{Dataset:} 
Our model is tested on the demanding JRDB-Act dataset~\cite{ehsanpour2022jrdb}, which captures dynamic social interactions in crowded indoor and outdoor environments on the Stanford University campus. Social group annotations are determined based on the behavior and activities of individuals within the scene. Adhering to the JRDB-Act protocol, the effectiveness of our model's social group detection capabilities is evaluated on key frames selected every 15 frames from the dataset. The dataset comprises a total of 1419 training samples, 404 validation samples, and 1802 testing samples. \\
% We evaluate our model on JRDB-Act dataset \cite{ehsanpour2022jrdb}  which is a challenging dataset, captured by a social mobile robot placed in various indoor and outdoor places in Stanford university campus, containing many crowded scenes. JRDB-Act provides social group annotation that is based on their interactions in the scene to infer common activities in each social group (social activity). Following the setting in JRDB-Act, we evaluate the social group detection task on key frames that are sampled every 16 frames. 
% \siminj{The number of training, validation and testing samples are 1419, 404 and 1802, respectively.}\\
\textbf{Evaluation Metric:} We employ the mean Average Precision (mAP) metric, introduced in~\cite{ehsanpour2022jrdb}, which calculates the Average Precision (AP) for various group sizes ranging from one to more than five members indicated by G1 (one member),
G2 (two members), and G5+ (more than five members) in Table~\ref{table:ValidPerforme}. The evaluation for each task is conducted using the corresponding ground truth labels, consistent with the methodology outlined in~\cite{ehsanpour2022jrdb}.  \\
% We use the mean Average Precision (mAP), introduced by~\cite{ehsanpour2022jrdb}, that reports the Average Precision for each group size varying from one to more than 5 members, which is then averaged to calculate mAP.
% Consistent with~\cite{ehsanpour2022jrdb}, evaluation for each task is performed by considering the corresponding ground truth labels.  \\
\textbf{Implementation Details:}
The proposed method employs a sequence of bounding boxes, consisting of a maximum of 15 frames selected from 30 frames prior to the key-frame by choosing every other frame. The frames are then fed to LSTM module with a feature dimension of 32. The graph transformer component of the model is comprised of 5 layers with 8 attention heads.

The function $F^{L_1}$, as defined in Eq. (\ref{Eq7}), is a (2$\times$1) multi-layer perceptron (MLP) layer with a sigmoid activation function. The functions $F^{L_2}$ and $F^{L_3}$, defined in the equations, are MLP layers with dimensions (32$\times$16) and (1$\times$16), respectively, and utilize a ReLU activation function. The function $F^{L_4}$, defined in Eq. (\ref{Eq2}), is a (32$\times$1) linear layer.

The optimization process is performed using the ADAM optimizer with hyperparameters  $\eta_1 = 0.9$, $\eta_2 = 0.999$, and $\epsilon=10^{-8}$. The parameters in Eq.( \ref{losses_eq}) are set to $\lambda_1=5\times 10^{-4}$, $\lambda_2=0.1$ and $\lambda_3=5\times 10^{-4}$ and $\alpha$ and $\beta$ in Eq. (\ref{l_eig}) are both equal to 1. The model is trained for 50 epochs using a mini-batch size of 1 and an initial learning rate of $10^{-4}$, which is reduced by a factor of $0.1$ if the validation loss plateaus.

The experiment is conducted on an NVIDIA GeForce GTX 1080 Ti and a comparison is made with previous state-of-the-art frameworks~\cite{ehsanpour2020joint,ehsanpour2022jrdb,han2022panoramic} in terms of inference time. For a fair comparison, only the feed-forward time of each neural architecture is reported in the results.
% % ------
% We consider 30 frames before the key-frame and choose every other frame. \simin{Therefore, each sequence of bounding boxes consists  maximum of 16 frames that are feed to LSTM module. We use one layer of LSTM with feature dimension of 32. Moreover, the number of layers in graph transformer is 5 with 8 attention heads.}  Furthermore, $F^{L_1}( \cdot; \Theta_{3})$ in Eq.~\ref{Eq7} is a (2$\times$1) MLP layer with a sigmoid activation function.
% Also,  $F^{L_2}( H_N ;\Theta_{4}) $ and  $F^{L_3}( D ;\Theta_{5})$ are MLP layers with dimension (32$\times$16) and (1$\times$16) with a Relu activation function, respectively and $F^{L_4}( \cdot ;\Theta_{6})$ in Eq.~\ref{fig:framework} is (32$\times$1) a linear layer.
%  We utilize ADAM optimizer with the hyper parameters $\eta_1 = 0.9$, $\eta_2 = 0.999$, $\epsilon=10^{-8}$. Also, the parameters in Eq. \ref{losses_eq}  are $\lambda_1=5\times 10^{-4}$, $\lambda_2=0.1$ and $\lambda_3=5\times 10^{-4}$ and $\alpha$ and $\beta$ in Eq.~\ref{l_eig} are set to be both equal to 1.
% We train the model with for 50 epochs with a mini-batch size of 1 and an initial learning rate of $10^{-4}$. The learning rate is reduced by a factor of $0.1$ on validation loss plateaus.
% \simin{We conduct our experiment for all methods on GeForce GTX 1080 Ti, NVIDIA Corporation and compare the inference time of our model with the previous state-of-the-art frameworks~\cite{ehsanpour2020joint,ehsanpour2022jrdb}. }  
% For a fair comparison between the inference time of the competing models in 
\begin{table*}[tbh] 
  \begin{center}
    {\footnotesize{
% \begin{tabular}{c|cccc||ccccc|c}
% \toprule
% Model &Det/Track & D_{E} & GAT/G.Trans & Res Co &G1 AP$\uparrow$ & G2 AP$\uparrow$ & G3 AP$\uparrow$ & G4 AP$\uparrow$ & G5$^+$ AP$\uparrow$ & mAP$\uparrow$ \\
% \midrule
% Ablation1 &Det & \xmark &GAT&\xmark& 20.5&38.1&52.4&60.2&70.0&48.2\\
% Ablation2 & Track & \xmark &GAT&\xmark& 32.8&57.7&67.1&74.6&67.3&59.8\\
% Ablation3 & Track & \cmark &GAT&\cmark& 46.8&67.6&74.0&71.2&50.6&62.0\\
% % Ablation2  & Track & \xmark & G.Trans &\cmark&72.7&71&62.2&65.7&44.2&63.2&68.2\\
% % Ablation3  & Det & \cmark &G.Trans&\cmark& 63.1&65.7&63.6&76.9&55&64.8&63.7\\
% % Ablation4  & Track & \cmark &G.Trans&\xmark& 68.5 &68.9&61.7&73.6&52.1&64.9&66.6\\
% \textbf{Our Final Model}& Track & \cmark & G.Trans &\cmark &71.9 & 73.6& 64.0&71.2&48.6&{\bf 65.9}\\

% \bottomrule
% \end{tabular}
\begin{tabular}{l|cccc|ccccc|c}
\toprule
Model&Det/Track & $D_{E}$ & GAT/G.Trans & Res Co &G1 AP$\uparrow$ & G2 AP$\uparrow$ & G3 AP$\uparrow$ & G4 AP$\uparrow$ & G5$^+$ AP$\uparrow$ & mAP$\uparrow$  \\
\midrule
GAT/Graph Transformer & Track & \cmark & GAT &\cmark &46.8&67.6&74.0&71.2&50.6&62.0\\
w/o Euclidean Distance& Track & \xmark & G.Trans &\cmark& 72.7&71&62.2&65.7&44.2&63.2\\
w/o Motion Trajectory & Det & \cmark & G.Trans &\cmark &63.1&65.7&63.6&76.9&55&64.8\\
w/o Residual Connection& Track & \cmark & G.Trans &\xmark &68.5 &68.9&61.7&73.6&52.1&64.9\\

{\bf Our Final Model}& Track & \cmark & G.Trans &\cmark &71.9 & 73.6& 64.0&71.2&48.6&{\bf 65.9}\\

\bottomrule
\end{tabular}
}}
\end{center}
\caption{An ablation study on JRDB-Act validation-set using the ground truth Detection/Tracks.}
\label{ablastion1}
\end{table*}
\subsection{Results}
Table~\ref{table:ValidPerforme} displays a comparison between our model and the other state-of-the-art methods on the both validation and test JRDB-Act data. 
Our proposed method has achieved considerable performance improvement compared to the state-of-the-art approaches, \ie ~\cite{ehsanpour2022jrdb},~\cite{han2022panoramic},~\cite{ehsanpour2020joint}, and~\cite{li2022self}, in the both validation and test data. %Specifically, we attained mAP score of 65.9\% in the validation data and 32.5\% for the test data. In comparison, achieved lower mAP scores for validation data with values of 59.9\%, 47.1\%, 41.4\%, and 23.3\%, respectively. 
%As for test data,~\cite{ehsanpour2020joint} and~\cite{ehsanpour2022jrdb} obtained mAP scores of 24.5\% and 31.6\%, respectively. 
Moreover, our approach attained either the highest or second-highest AP scores across various group size categories in both the validation and test data. However, our method did not perform well in identifying one-member groups (G1) due to the difficulty in detecting individuals in crowded scenes where they are in close proximity to one another. In such situations, distance-based techniques may not provide accurate results. \\
% Our proposed method has achieved considerable improvement with mAP score of 65.9\% and 32.5\% for validation and test data, respectively. While other approaches~\cite{ehsanpour2022jrdb},~\cite{han2022panoramic} ,~\cite{ehsanpour2020joint} and~\cite{li2022self}  have achieved mAPs of 59.9\%, 47.1\%, 41.4\% and 23.3\% for validation data and  for test data,~\cite{ehsanpour2020joint}  and ~\cite{ehsanpour2022jrdb} obtained 24.5\% and 31.6\%, respectively. 
% including~\cite{ehsanpour2020joint} and~\cite{ehsanpour2022jrdb}, achieving an mAP of , which is better than their scores of 24.5\% and 31.6\%, respectively.
%n contrast,~\cite{li2022self} have exhibited low performance with mAP of 23.3\%. 
One of the main reasons for the success of our approach is its reliance on simple and efficient input data (bounding boxes) along with the encoding of each individual's trajectory over time. This approach is less prone to noise compared to the other competitive methods using more complex input data, \eg video or body pose, while offering significant computational efficiency compared to them.
% such as body pose, as demonstrated in [21].
% Moreover, other appearance-based methods such as~\cite{ehsanpour2020joint},~\cite{ehsanpour2022jrdb},~\cite{han2022panoramic}, which rely on observations using I3D or Inception3 are
% prone to occlusions or noisy observations. 
Furthermore, we employ a graph transformer to model the relationships between nodes, which may prove more effective. By incorporating trajectory information into node features and geometric distances between subjects in edge features, we can achieve more accurate modelling of social relations. 
% Our approach, graph transformer, has several advantages over GAT including scalability, simplicity, generalizability, and flexibility that has been discussed in Ablation Section.
% than GAT as used in previous studies~\cite{ehsanpour2020joint}~\cite{ehsanpour2022jrdb}, or other graph-based models like the hierarchical graph neural network~\cite{han2022panoramic}. By incorporating trajectory information into node features and geometric distances between subjects in edge features, we can achieve more accurate modelling of social relations. In contrast, GAT may struggle to capture such interactions, as it relies on simple pairwise attention mechanisms. Our approach, graph transformer, has several advantages over GAT, including scalability, simplicity, generalizability, and flexibility that has been discussed in Ablation Section.
Moreover, in our approach, the distances between trajectories are employed as edge features to serve as priors in the graph representation, which differs from the approach taken by~\cite{ehsanpour2020joint} and~\cite{ehsanpour2022jrdb} where the graph is fully connected as an input to the GAT.\\
Note that, all the methods reported in Table~\ref{table:ValidPerforme} used ground truth bounding boxes for reporting their results on the validation set. However, since no ground-truth bounding box was available for the test data, the performance of all the methods on this set relied on the detections/tracks generated by~\cite{he2021know}. Although its detection performance is reasonable (i.e. AP0.5 = 68.1\%), its tracking performance is not as well as its detection and is moderately low, (i.e. MOTA = 32.3\%).
% It is worth noting that, with the exception of method~\cite{li2022self}, which employs self-supervised techniques, all other methods used ground truth labels for validation evaluation. Additionally, since no ground truth bounding box was available for the test data, all methods relied on the same algorithm~\cite{he2021know} for detecting bounding boxes, which can also be used for tracking purposes. However, the performance of ~\cite{he2021know} is unsatisfactory, as indicated by its accuracy (MOTA) of approximately 32.3\%, which is lower than its detection accuracy (AP0.5 = 68.1\%). 
We anticipated this reduction in performance for test data and believe that a better tracking algorithm could further improve our method's results. It is important to mention that we have presented results for~\cite{ehsanpour2020joint} and~\cite{ehsanpour2022jrdb}, while the test results for~\cite{han2022panoramic} and~\cite{li2022self} were unavailable.\\
 Table~\ref{table:testPerforme} displays the number of parameters used by each model (Num. Parameters) and its processing speed, represented in frames per second (fps).
In terms of the number of parameters, model~\cite{li2022self} has a relatively small number of parameters (1.6 million), while models~\cite{ehsanpour2020joint,ehsanpour2022jrdb}, and~\cite{han2022panoramic} have significantly larger numbers of parameters, at 27 million, over 27 million (indicated by 27M$^{+}$), and 23 million, respectively. This increase in parameters is due to the use of modules such as  I3D, RoiAlign, and GAT. Our method, which has 700,000 parameters and LSTM and graph transformer as its main modules, represents a trade-off between model complexity, capacity to learn from data, and computational efficiency.
\siminj{
In terms of computational cost (fps), model~\cite{li2022self} has a fps rate of less than 29.41 (indicated by 29.41$^{-}$) due to its number of parameters. Models~\cite{ehsanpour2020joint,ehsanpour2022jrdb} have lower fps rates of less than 2.40 and 2.3, respectively, due to their high number of parameters. Model~\cite{han2022panoramic} also has a low fps rate of less than 2.3 due to its use of the same backbone as~\cite{ehsanpour2020joint,ehsanpour2022jrdb} and GCN modules. Our method, with 700,000 parameters and LSTM and graph transformer as its main modules, has a fps rate of 29.41, which is the best, highlighting the trade-off between model complexity, capacity to learn from data, and computational efficiency.}\\
\\
\textbf{Qualitative Results:} 
We visualize the detected groups in selected images from the validation set in Fig.~\ref{fig:vis1}. In each pair, images on the left show the ground truth $(a1)$, $(b1)$ and $(c1)$, while the images on the right illustrate the output of our model $(a2)$, $(b2)$ and $(c2)$.
From the comparison, we can observe that our model performs well not only in simple scenes such as $(a2)$ but also in challenging scenes such as $(b2)$. In $(a2)$, GP1, GP2, and GP3 are evidence of our model's excellent performance, corresponding to GT1, GT2, and GT3 in the ground truth image $(a1)$, respectively.
However, our model still faces difficulties in detecting challenging groups where individuals are far apart in depth but close in x, y coordinates. For instance, in the ground truth image $(b1)$, GT2 and GT3 are separate groups, but they are considered as one group, GP2, in  $(b2)$. Similarly, in $(c1)$, we can see that GT2 and GT3 are separated, but our model detects them as one group, GP2. This issue arises when, in some scenes, the trajectory of each individual is not well encoded and the distance calculation technique considers them as a group. 
To overcome this limitation, incorporating more information and encoding a richer feature representation could be beneficial in detecting interacting people in social groups. For example, considering interaction, body pose, and head orientation in addition to calculating distances could greatly enhance the social group detection capabilities of our model.
\subsection{Ablation Study}
\siminj{
The efficacy of each constituent element in our proposed methodology is demonstrated through a comprehensive ablation analysis in Table ~\ref{ablastion1}. The final performance on the JRDB-Act validation set is evaluated by systematically removing each essential component, allowing us to assess their individual impact on the final performance. These experiments are conducted using ground truth bounding boxes from the validation set to eliminate the influence of detection performance.
One key element of the ablation is the evaluation of the graph transformer, as depicted in Fig.\ref{fig:framework}. To evaluate its effectiveness, the GAT model \cite{velivckovic2017graph} is adopted instead of the graph transformer (G.Trans), as indicated by the "GAT/G.Trans" entry in Table ~\ref{ablastion1}. Additionally, the Euclidean distance ($D_{E}$) as defined in Eq. (\ref{De}) is used to calculate the distance between the updated node features ($H_{N}$).
The role of trajectory is also explored, as depicted by the "Det/Track". The "Det" notation indicates that only a single key-frame is considered, with no trajectory information provided to the LSTM module, while the "Track" notation indicates that 15 frames are considered, allowing us to calculate the distance between trajectories as described in Sec.III "Calculating a Distance Between Two Variable-sized Tracks". Lastly, the residual connection depicted in Fig.\ref{fig:framework}, which connects the input of the graph transformer to its output, is evaluated and referred to as "Res Co" in Table ~\ref{ablastion1}. Further details regarding each ablation can be found in the following:}
\siminj{
\textbf{GAT vs Graph Transformer}: In this ablation study, the GAT model is compared against the graph transformer. The other components are kept unchanged. As displayed in Table~\ref{ablastion1}, the results show that the performance decreases by 5.9\% mAP when using the GAT model instead of the graph transformer. This disparity can be attributed to the multi-headed attention mechanism employed by the graph transformer, which significantly improves attention compared to the GAT model. Furthermore, as demonstrated in~\cite{dwivedi2020Generalization}, the generalization performance of transformer networks to graph data is significantly higher than that of GAT models.}
\siminj{
\textbf{With vs Without Euclidean Distance}: In this ablation, the contribution of the Euclidean distance ($D_{E}$) is evaluated. The $D_{E}$, which calculates the Euclidean distance between the updated node features, is removed, and instead, the $H_{N}$ is concatenated with D as described in Eq.(\ref{Eq6}). This change transforms Eq.(\ref{Eq6}) into $H_{c}=H_{N}\mathbin\Vert D$. The Euclidean distance between the encoded motion trajectory, updated by the graph transformer, is combined with the pairwise track distance, contributing to a more effective encoding of the distance between individuals in a group. Table~\ref{ablastion1} shows a 4\% decrease in mAP.}
% \siminj{\textbf{GAT vs Graph Transformer}: In this ablation, the GAT is used instead of graph transformer and other components remains unchanged. As it can be seen in Table 3, using GAT, the performance decreased by 5.9\% mAP. 
% % improved by 6.29\% mAP1.
% The reason is that adopted graph transformer employs multi-headed attention that noticeably improved attention. Furthermore, proved by \cite{dwivedi2020generalization}, the performance of generalization of transformer networks to graphs is significantly higher than GAT. \\
% \textbf{With vs Without Euclidean Distance}: The contribution of euclidean distance is depicted in this ablation, $D_{E}$, the euclidean distance between updated node features,
% is removed and instead $H_{N}$ is concatenated with $D$, in Eq \ref{Eq6}. In other words, Eq \ref{Eq6} changes to $H_{c}=H_{N}\mathbin\Vert D$. Euclidean distance between encoded motion trajectory, updated by graph transformer, is aggregated with the pairwise track distance, contributing to better encoding distance between individuals in a group. Shown in Table 3, the effect of this ablation is mainly on mAP by 4\%.\\
\\
\textbf{With vs Without Motion Trajectory}: This ablation study evaluates the impact of removing the motion trajectory of individuals in our model. The 15 frames are eliminated, and only the last frame is used to calculate the distance,  ${d}_{i,j}$. As a result, instead of calculating the distance between two trajectories, $\bm{d}_{i,j}(P_{i},P_{j})$, the distance between two bounding boxes is calculated. Consequently, only one frame is fed into the LSTM, while the rest of the model remains unchanged. The results show that considering the motion trajectory improves the performance by 1.69\% in terms of mAP.\\
\textbf{With vs Without Residual Connection}: The architecture of our proposed method, as shown in Fig.~\ref{fig:framework}, includes a residual connection that connects the input of the graph transformer to its output. This ablation study aims to verify the role of the residual connection in our model by removing it. The results, shown in Table~\ref{ablastion1}, indicate that the residual connection outperforms the model without it by a margin of 1.5\%.
It is believed that the residual connection helps the graph transformer to more efficiently learn the distance between instances, thus allowing the model to learn better.

\section{CONCLUSIONS}
In this paper, we present a simple, but efficient, end-to-end supervised model that represents the first graph transformer-based approach in the field of social group detection. Our solution stands out in its ability to use only the coordinates of bounding boxes as the input to the model, while simultaneously achieving faster performance than previous state-of-the-art models. We model the social group detection problem as a graph, with the node features estimated by encoding tracks using LSTM and the edge values determined by the time-average distance between each track pair. Our model achieves state-of-the-art results on the JRDB-Act benchmark dataset and outperforms previous methods in terms of both performance and speed. 
%Additionally, our model is highly efficient, boasting a small number of parameters and exceptional computational speed. 
In future, we aim to enrich the input feature representation for this task by incorporating additional information such as individuals' actions, body poses, and head orientations. This will lead to an even richer and more comprehensive representation of the social group dynamics.

\addtolength{\textheight}{-12cm}   % This command serves to balance the column lengths

%%%%%%%%%%%%%%%%%%%%%%%%%%%%%%%%%%%%%%%%%%%%%%%%%%%%%%%%%%%%%%%%%%%%%%%%%%%%%%%%

%%%%%%%%%%%%%%%%%%%%%%%%%%%%%%%%%%%%%%%%%%%%%%%%%%%%%%%%%%%%%%%%%%%%%%%%%%%%%%%%



% \section*{ACKNOWLEDGMENT}

% The preferred spelling of the word ÒacknowledgmentÓ in America is without an ÒeÓ after the ÒgÓ. Avoid the stilted expression, ÒOne of us (R. B. G.) thanks . . .Ó  Instead, try ÒR. B. G. thanksÓ. Put sponsor acknowledgments in the unnumbered footnote on the first page.

%%%%%%%%%%%%%%%%%%%%%%%%%%%%%%%%%%%%%%%%%%%%%%%%%%%%%%%%%%%%%%%%%%%%%%%%%%%%%%%%
%\bibliographystyle{Simin_bib} 
\bibliography{Simin_bib.bib}

\end{document}